%% file: main-lrec-coling2024.tex
\documentclass[10pt, a4paper]{article}

\usepackage[final]{lrec-coling2024} 

\usepackage{subfig}
\usepackage{amsmath}
\usepackage{amssymb}
\usepackage{amsfonts}
\usepackage[ruled,linesnumbered]{algorithm2e}
\usepackage{booktabs}    
\usepackage{graphicx} 
\usepackage{multicol}
\usepackage{multirow}

\title{FlattenQuant: Breaking Through the Inference Compute-bound for Large Language Models with Per-tensor Quantization}
\name{Yi Zhang,  Fei Yang$^{\ast}$ \thanks{*Fei Yang is the corresponding author.}, Shuang Peng, Fangyu Wang, Aimin Pan} 

\address{Zhejiang Lab, Hangzhou, China \\        
         \{zhangyi620, yangf, pengs, wangfy, panaimin\}@zhejianglab.com\\
         }

\abstract{Large language models (LLMs) have demonstrated state-of-the-art performance across
various tasks. However, the latency of inference and the large GPU memory consumption of LLMs restrict their deployment performance. Recently, there have been some efficient attempts to quantize LLMs, yet inference with large batch size or long sequence still has the issue of being compute-bound. Fine-grained quantization methods have showcased their proficiency in achieving low-bit quantization for LLMs, while requiring FP16 data type for linear layer computations, which is time-consuming when dealing with large batch size or long sequence. In this paper, we introduce a method called FlattenQuant, which significantly reduces the maximum value of the tensor by flattening the large channels in the tensor, to achieve low bit per-tensor quantization with minimal accuracy loss. Our experiments show that FlattenQuant can directly use 4 bits to achieve 48.29\% of the linear layer calculation in LLMs, with the remaining layers using 8 bits. The 4-bit matrix multiplication introduced in the FlattenQuant method can effectively address the compute-bound caused by large matrix calculation. Our work achieves up to 2$\times$ speedup and 2.3$\times$ memory reduction for LLMs with negligible loss in accuracy. 
 \\ \newline  }

\begin{document}
\maketitleabstract
\section{Introduction}
\input{Intro.tex}

\section{Related Work}
\input{Relate.tex}

\section{Bottleneck of LLMs Quantification}
\input{Bottleneck.tex}

\section{FlattenQuant}
\input{FlattenQuant.tex}

\section{Experiments}
\input{Experiment.tex}

\section{Conclusion and Future Work}
\input{conclusion.tex}

\section*{Acknowledgement}
This work is supported by National Natural Science Foundation of China (No.U22A6001), Key Research Project of Zhejiang Lab (No.2022PG0AC02,  No.K2023NB0AC11), Key R \&D Program of Zhejiang (No.2022C04006), and Natural Science Foundation of Zhejiang Province (No.LQ21F020004).

\nocite{*}

\section{References}
\bibliographystyle{lrec-coling2024-natbib}
\bibliography{Reference.bib}

\bibliographystylelanguageresource{lrec-coling2024-natbib}
\bibliographylanguageresource{languageresource}

\end{document}

%% file: Intro.tex
\begin{figure}[!t]
\begin{center}
\includegraphics[scale=0.5]{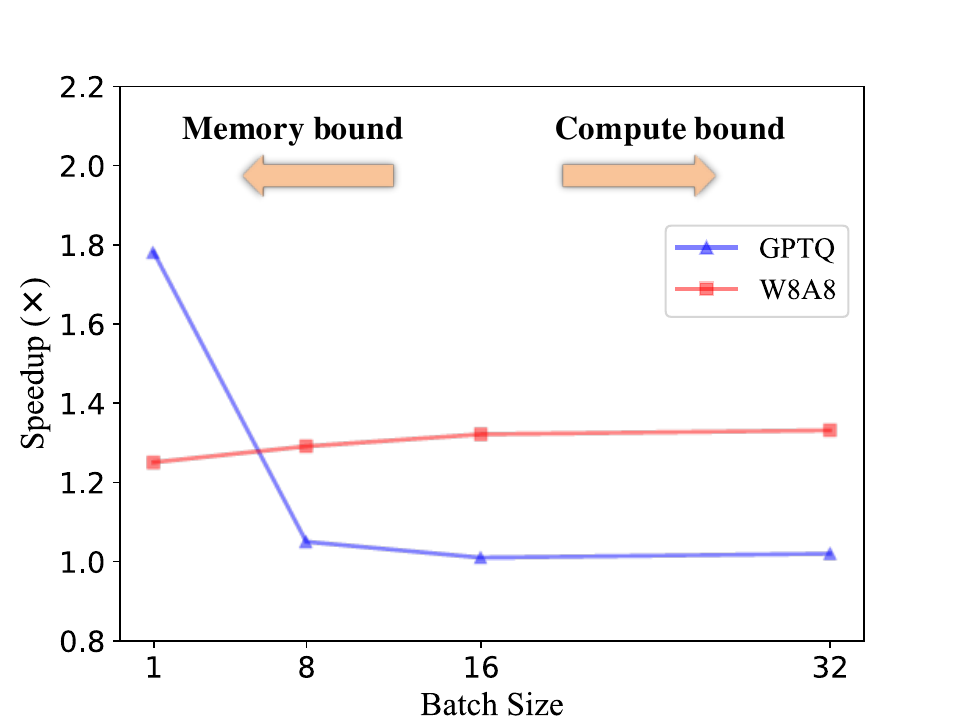}
\caption{As the batch size increases, the determining factor for inference latency shifts from being memory-bound to compute-bound. GPTQ prioritizes memory optimization while utilizing FP16 computation, whereas W8A8's 8-bit computation prioritizes computation optimization. }
\label{GPTQ_W8A8}
\end{center}
\end{figure}

The impressive capabilities of large language models (LLMs) have made a significant impact in recent years \cite{OpenAI_2023,ge2023openagi,zhao2023survey}. Various LLMs have been released and applied in the real-world production environment \cite{eloundou2023gpts}. As a result, there is a widespread need for the deployment of LLMs.\\
Using LLMs for inference results in a significant consumption of hardware memory resources due to the large number of weight parameters and activation tensor caches generated. Furthermore, inference of transformer layers necessitates intensive matrix calculations, posing a significant challenge to GPU computational capabilities. The facts mentioned above result in a memory-bound and a compute-bound for LLM inference, respectively, which lead to significant inference delay.\\
With the burgeoning use of LLMs, there is an escalating need to efficiently process a significant influx of inference requests simultaneously, necessitating the execution of inference using large batch sizes. A widely adopted method to optimize LLM inference is GPTQ quantization, as presented in \cite{frantar2022gptq}, employing 4-bit quantization for weights. This effectively mitigates memory-bound issues, especially with small batch size or short sequence, resulting in impressive performance. However, GPTQ does not extend quantization to activations, still relying on FP16 for computations instead of transitioning to lower bit levels. Consequently, it faces compute-bound challenges as batch size or sequence length increases. Figure \ref{GPTQ_W8A8} visually demonstrates this challenge, where 8-bit computation shows superior acceleration, particularly evident with a sequence length of 256, compared to GPTQ utilizing FP16 computation. This phenomenon is rooted in the compute-bound nature of the inference process, where latency is predominantly influenced by matrix computations rather than memory access.\\
In small-scale models such as CNNs, 8-bit quantization can ensure a small loss of accuracy and effectively reduce inference delay \cite{2019Post}. However, LLMs present two challenging aspects when it comes to quantification:
\begin{itemize}
\item In the inference process of LLMs, there is a compute-bound problem when dealing with large batch size or long sequence. To speed up matrix calculations, there is a practical demand for using 8 bits or even 4 bits.
\item Due to the significant difference in the value distribution between each channel, most existing schemes use per-channel quantization \cite{frantar2022gptq} or group-wise quantization \cite{yao2022zeroquant}. However, these fine-grained quantization approaches uses different scaling factors inside the tensor, which can prevent the calculation of the linear layer from using low-bit matrix multiplication directly, thus slowing down the inference speed.
\end{itemize}
In this paper, we propose a method called FlattenQuant to achieve accurate and low-bit per-tensor quantization. This method involves flattening channels with large values and adding extra channels to accommodate these values. Through this process, the maximum value of the tensor is considerably reduced while preserving complete information. The flattened tensor can undergo per-tensor quantization while ensuring the preservation of accuracy.
In summary, our contributions are as follows:
\begin{itemize}    
    \item We analyze the relationship between quantization schemes of LLMs and inference latency, especially the requirement for quantization schemes to overcome the compute-bound.
    \item Based on the findings, we propose FlattenQuant, which reduces the difficulty of per-tensor quantization and enables low bit computation to overcome compute-bound.
    \item We propose a quantization framework based on FlattenQuant, which can directly use 4 bits to achieve 48.29\% linear layer computation in LLMs, and 8 bits for the remaining. Compared to baselines computed using FP16, we achieve up to 2$\times$ speedup and 2.3$\times$ memory reduction, with only a very minor accuracy loss.
\end{itemize}

%% file: Relate.tex
\subsection{Neural Network Quantization} 
Quantization is an efficient method for model compression during the deployment \cite{jacob2018quantization}.
Due to the high hardware requirements of LLMs training, we focus on deployment-friendly post-training quantization (PTQ) \cite{hubara2021accurate}.\\
Initiating the PTQ process requires a calibration dataset, which is essential for collecting the numerical distribution of input activation for each layer of the network. With this dataset, we can derive the required quantization parameters for each layer by analyzing the value distribution of activation and weight. The most commonly employed method for obtaining quantization parameters can be expressed using the following equation:

\begin{equation}
\label{eq:quant}
X_{k}= \mathrm{round}(\frac{X_{fp16}}{s}),\quad s=\frac{\max(\left|{X_{fp16}}\right|)}{2^{k-1}-1}\enskip
\end{equation}
In the quantization process (illustrated in Equation \ref{eq:quant}), the floating-point activations of the layer input are denoted as $X_{fp16}$, while the quantized integer values are represented as $X_{k}$. The desired bit width is indicated by $k$, and the scaling factor from floating-point to integer is denoted by $s$. The final quantization accuracy is significantly influenced by the selection of scaling factor, which is crucial to the process.


\subsection{Per-Tensor and Fine-Grained Quantization} 
Quantifying tensors can be accomplished in various ways, each offering different levels of granularity. When a scaling factor is for the entire tensor, it is referred to as per-tensor quantization. Alternatively, we can employ alternative methods to refine the quantization granularity, such as per-token \cite{yao2022zeroquant}, per-channel \cite{frantar2022gptq}, or group-wise \cite{yao2022zeroquant,2021VS} quantization. In practice, for Convolutional Neural Networks (CNNs), per-tensor quantization is the most commonly used method because it can meet accuracy requirements, involves fewer quantization parameters, and is less computationally demanding during inference. However, achieving the necessary accuracy through direct per-tensor quantization can be challenging due to the complex numerical distribution of LLMs. Consequently, numerous fine-grained quantization methods have been explored based on data distribution characteristics.

%% file: Bottleneck.tex
\subsection{Outliers of Activation Tensors}
Research has shown that LLMs typically possess uniformly distributed weights, simplifying quantification through the per-tensor method \cite{xiao2023smoothquant}. However, quantifying activations poses a challenge due to the varying value distribution across channels \cite{xiao2023smoothquant}. Some channels exhibit values 20-100 times larger than the majority, necessitating the truncation of maximum values to establish an appropriate threshold for scaling factors. If the threshold is set too small, vital information from these larger activation channels may be lost. Conversely, if the threshold is too large, the quantization accuracy of most channels will be significantly reduced. In such situations, quantization for LLMs often adopts finer-grained methods. These include employing varying bit widths for different tensor segments, per-channel quantization, or group-wise quantization based on numerical distribution.\\
\begin{table}[!t]
\centering
\caption{Setting of the LLMs quantification}
\label{DiffQuantizationSettings}
\resizebox{\linewidth}{!}
{
\begin{tabular}{cccc}
\toprule
Method & Activation & Weight & Compute data type\\
\midrule
W8A8 &per-tensor static &per-tensor & INT8\\ 
LLM.int8() &per-token dynamic &per-channel & INT8+FP16\\ 
SmoothQuant &per-tensor dynamic/static&per-tensor & INT8\\ 
RPTQ &group-wise static& group-wise & FP16\\ 
GPTQ &not quant &per-channel & FP16\\ 
\bottomrule
\end{tabular}
}
\end{table}
%
As shown in Table \ref{DiffQuantizationSettings}, various quantization approaches have been proposed for handling LLMs. The LLM.int8() approach \cite{2022LLM} leverages the FP16 data type to handle the tensor that are difficult to quantize and utilizes the INT8 data type for the remainder. GPTQ \cite{frantar2022gptq} focuses on 4 or 3-bit quantization of weights, utilizing per-channel quantization and adjusting unquantized parameters based on the Hessian matrix. RPTQ \cite{yuan2023rptq} clusters the numerical distribution of activation tensors and groups them accordingly for low-bit quantization, albeit at the expense of potential inference efficiency reduction due to memory rearrangement. In contrast, SmoothQuant \cite{xiao2023smoothquant} seeks to balance the difficulty of activation and weight quantization by transferring large values from the activation tensor to the weight. However, this may not be effective if the activation tensor contains excessively large outliers.

\subsection{Aim for Compute-bound}
Inference latency is influenced by two primary factors: compute-bound and memory-bound. Quantization serves to alleviate the memory bottleneck significantly. For example, when GPTQ quantizes the weights of LLMs to 3 bits, it results in over a 3$\times$ acceleration in inference on A100 GPUs. However, as the input batch size and the sequence length increase, the compute-bound factor becomes predominant, overshadowing the influence of memory-bound. In such cases, matrix multiplication consumes up to 80\% of the inference time, as reported by LightSeq \cite{wang2020lightseq}. Consequently, the primary approach to mitigate the compute-bound challenge is to reduce the time required for matrix multiplication.\\
Taking the A100 \cite{choquette20213} GPU as an example, in terms of computing power, INT4 computation demonstrates a 4$\times$ acceleration compared to FP16 computation, while INT8 showcases a 2$\times$ improvement. The aforementioned statement highlights the possibility of decreasing the bit width in order to tackle compute-bound difficulties in the context of large-scale matrix multiplication.\\
When utilizing the fine-grained quantization method, a challenge arises concerning the compatibility of quantization units and matrix multiplication calculations. As a result, direct utilization of TensorCore \cite{markidis2018nvidia} for performing matrix multiplication on quantized activation and weight becomes unfeasible. For instance, in the case of methods like LLM.int8() \cite{2022LLM}, which is imperative to partition the matrix multiplication calculation into separate precision-based computations, followed by the summation of results. Techniques employing per-channel quantization \cite{frantar2022gptq} necessitate the dequantization of tensors to the FP16 data type before matrix multiplication calculations. Similarly, methods such as RPTQ \cite{yuan2023rptq}, which rely on group-wise quantization, faces challenge when performing linear layer calculations within a single matrix multiplication operation. These above methods have limited the quantization granularity of weights to the level of channels or even groups, indicating that each data unit in the tensor corresponds to different quantization coefficients. This makes the quantized weights unable to participate in matrix multiplication calculations directly, which necessarily requires dequantization to FP16 for computation. This will cause compute-bound under large batch sizes or long sequences. \textbf{Consequently, when compute-bound scenarios are anticipated, per-tensor quantization remains the preferable choice for deploying LLMs.}

%% file: FlattenQuant.tex
\begin{figure*}[!ht]
\begin{center}
    \subfloat[\label{1a}]
    {\includegraphics[width=1.55\columnwidth]
    {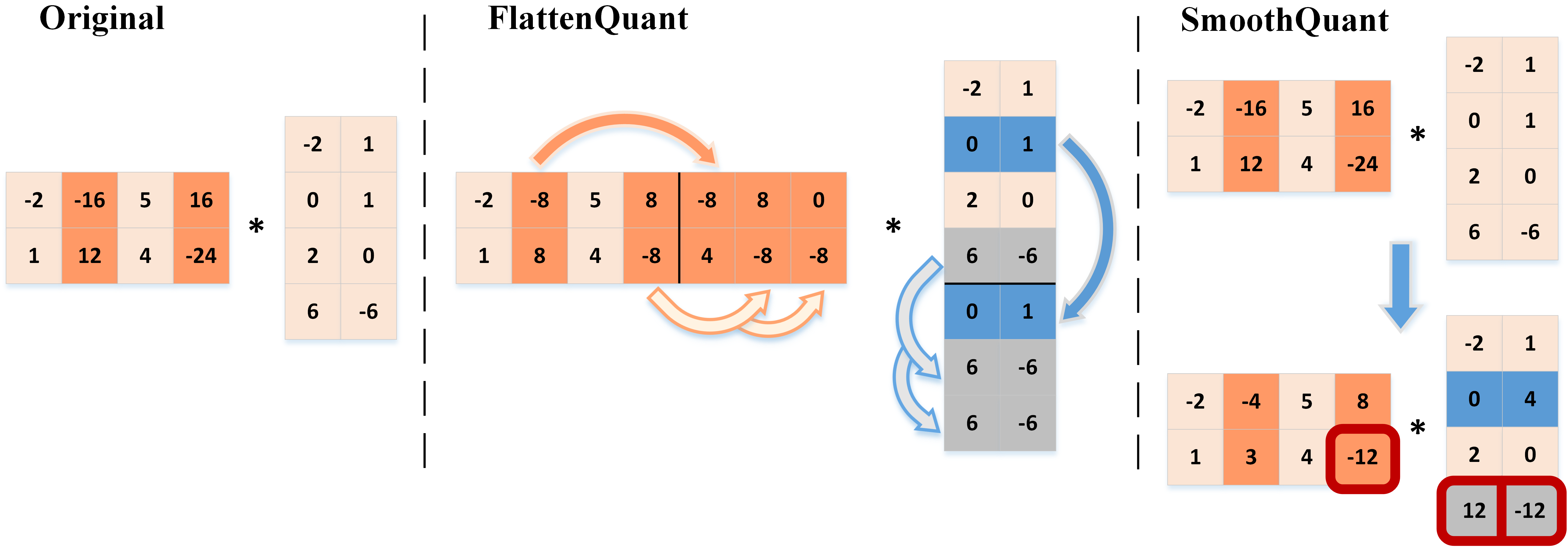}}
    \vspace{1pt}
    \subfloat[\label{1b}]
    {\includegraphics[width=1.75\columnwidth]{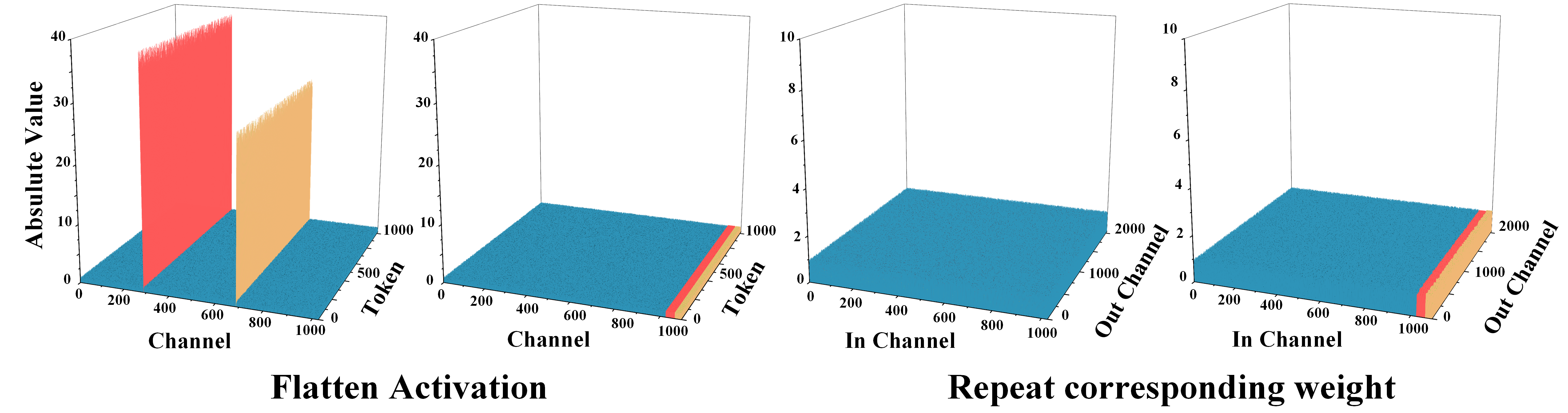}}
 
\caption{The concept of FlattenQuant is illustrated in (a). The expansion of certain channels is undertaken to accommodate those with larger values, another tensor repeats corresponding channels by the number of extended channels. Such operation itself does not result in any precision loss. We can observe that the activation's maximum absolute value in FlattenQuant is two thirds of that in SmoothQuant, accompanied by the weight's maximum absolute value being half of that in SmoothQuant. This flatter condition leads to a remarkable enhancement in quantization accuracy.
(b) displays the activation input and weight in LLM. After the flatten operation, the tensor becomes flat and easily per-tensor quantized.}
\label{fig1}
\end{center}
\end{figure*}

\subsection{Flattening the Tensor}
Our proposed approach, FlattenQuant, leverages per-tensor quantization to facilitate efficient low-bit matrix multiplication during linear layer computation in LLMs inference. By utilizing TensorCore, we ensure optimal performance. The core aspect of our approach involves identifying the channel indexes containing outliers in the target tensor, expanding these specific channels to accommodate the outliers, and repeating the corresponding matrix channels accordingly to ensure accurate matrix multiplication. This strategy is underpinned by the consistent presence of outliers between channels and the limited variance within each channel, as elucidated in \cite{bondarenko2021understanding}.\\
Preparing for the FlattenQuant, we employ a calibration dataset to perform inference on the model and identify the maximum value for each activation channel. Utilizing the numerical distribution, we establish a truncation threshold for the entire activation. Channels exceeding this threshold in their maximum value underwent flattening and were assigned new channels. The number of new channels are determined based on the maximum value of the original channel. This process is quantified by Equation \ref{flattenchannels}, which calculates the count of extended channels ($E_{j}$) for the $j$-th input channel, with $j$ ranging from $1$ to $C$ (the total number of input channels), and $T$ represents truncation threshold. The aforementioned approach results in a total count of extended channels, denoted as $C_{extend}$.

\begin{equation}
\label{flattenchannels}
E_{j}= \lfloor \frac{\max(\left|X_j\right|)}{T} \rfloor,
\qquad C_{extend}= \sum_{j=1}^{C}E_j
\end{equation}
Referring to Figure \ref{1a}, the activation channel is extended, and the weight channel is repeated accordingly. Equation \ref{Fatten} explains how the activation element $X_{ij}$ is flattened, and Equation \ref{Repeat} explains how the j-th channel of weight $W$ is repeated.
\begin{equation}\
\label{Fatten}
\begin{split}
\mathrm{Flatten}(X_{ij})=\overrightarrow{X_{i\widetilde{j}}}=[\overbrace{ T,\ldots,T }^{\lfloor\frac{X_{ij}}{T} \rfloor},X_{ij}\bmod T]\\
\widetilde{j}\in[j\cup(C+\sum_{k=1}^{j-1}E_k,C+\sum_{k=1}^{j-1}E_k+\lfloor\frac{X_{ij}}{T} \rfloor]]
\end{split}
\end{equation}

\begin{equation}
\label{Repeat}
\mathrm{Repeat}(W_j)={E_j}\left\{
\begin{bmatrix}
    1\\
    1\\
    \vdots\\    
\end{bmatrix}
\right.W_j
\end{equation}
%
%
%
%
Restricting the activation maximum value to a lower truncation threshold yields enhanced accuracy in the context of per-tensor quantization. Additionally, to further enhance the accuracy of per-tensor quantization, we also apply the flatten operation to the weights, which effectively decreases the maximum values of the weights and promotes a more uniform distribution of values.
\subsection{Achieving High-precision}

\begin{figure*}[!ht]
\begin{center}
\includegraphics[width=1.65\columnwidth]{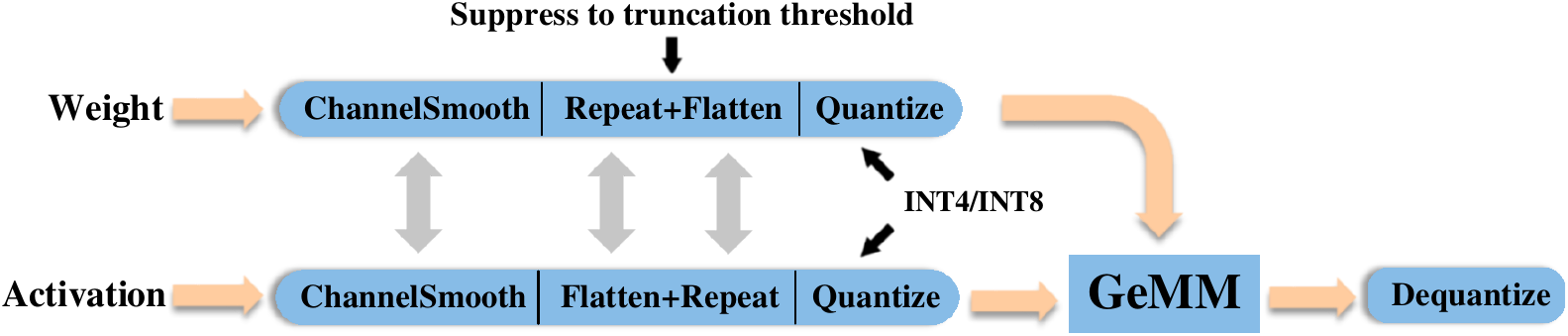}
 
\caption{Quantization framework based on FlattenQuant. The data flows through orange arrows, while gray arrows indicate operation configuration dependencies. Activation operations are executed during inference; and weight processing is done during model quantization.}
\label{overflow}
\end{center}
\end{figure*}

\paragraph{Smoothing across channels.}
To enable per-tensor quantization, SmoothQuant utilizes a smooth operation that relocates high-value activation data into the weights. We utilize a similar, albeit slightly modified, operation, as depicted in Equation \ref{smoothequation}, where $\alpha$ is the migration strength. We define $\mu$ as $\sum_{k=1}^{C}\max(\left|X_k\right|)/C$, where $\sigma$ is $(\sum_{k=1}^{C}(\max(\left|X_k\right|)-\mu)^2/C)^{0.5}$. In contrast SmoothQuant, our objective is to acquire uniformly distributed of channel values. Upon performing normalizing on the maximum absolute value of each channel, we obtain the relative numerical magnitude within the tensor and apply a smoothing process between activations and weights. This promotes a more equitable distribution of channel values, leading to flatter activations and weights.
%
\begin{center}
\begin{equation}
\label{smoothequation}
\begin{aligned}
&\mathrm{Smooth}(X,W)=(X \mathrm{diag}(s)^{-1},\mathrm{diag}(s)W)\\
&s_j=\frac{(\mathrm{Sigmoid}((\max(\left|X_j\right|)-\mu)/\sigma))^{\alpha}}{(\mathrm{Sigmoid}((\max(\left|W_j\right|)-\mu)/\sigma))^{1-{\alpha}}}
\end{aligned}
\end{equation}
\end{center}


\paragraph{Selection of the truncation threshold.}
The truncation threshold determines the maximum value before per-tensor quantization. A smaller threshold leads to higher precision in quantization, but also results in increased GPU memory consumption and linear layer computation. Our primary objective in selecting the threshold is to prevent outlier channels from interfering with quantization scaling factor and to avoid excessive flattening of channels. We start by using a boxplot \cite{frigge1989some} to suppress outlier channels. In Equation \ref{plotbox}, the lower quartile ($Q_1$) and upper quartile ($Q_3$) divide the data into the lowest and highest 25\% respectively, while the interquartile range ($IQR$) is defined as $Q_3-Q_1$. We then derive the truncation threshold (Equation \ref{TruncationThreshold}) by multiplying a coefficient $\beta$ with the mean of the maximum values of the channels.

\begin{equation}
\label{plotbox}
\begin{aligned}
\left|\widetilde{X_j}\right|=\mathrm{clip}(\left|X_j\right|,&\min=Q_1-1.5\cdot{IQR},\\&\max=Q_3+1.5\cdot{IQR})
\end{aligned}
\end{equation}

\begin{equation}
\label{TruncationThreshold}
T=\beta\cdot\frac{\sum_{j=1}^{C}\max(\widetilde{\left|X_j\right|})}{C}
\end{equation}

\paragraph{Quantize some layers to INT4.}
Smoothing across channels is instrumental in achieving a more uniform distribution of values across tensor channels, further flattening of the tensor, greatly suppressing the maximum value of the tensor, and significantly reducing the difficulty of quantization. The above operations serves as a critical prerequisite for 4-bit per-tensor quantization. Notably, the INT4 data type offers a representation range that is only $1/16$ of that of INT8. As matrix multiplication is performed after both activations and weights have been quantized to INT4, even slight quantization errors can have substantial impacts. Hence, it is imperative to attain high levels of quantization accuracy when employing INT4.\\
Our primary goal is to enable 4-bit quantization for specific linear layers, necessitating the assessment of each layer's suitability for allocation to 4-bit precision. To accomplish this, we have adopted a methodology akin to that utilized in TensorRT \cite{vanholder2016efficient} to evaluate the quantization-induced error, leveraging the KL divergence as a key metric for assessment. Upon performing per-tensor quantization to 4 bits and 8 bits, as outlined in Equation \ref{QuantBits}, we obtained the KL divergence between the quantized tensors distribution $Q$ and the original data distribution $P$. The resulting ratio was then compared against a predefined threshold $\gamma$. If the ratio falls below this threshold, the layer is assigned to 4-bit quantization; otherwise, it is allocated to 8-bit quantization. It is important to note that for a layer to be quantized to 4 bits, both activations and weights must undergo 4-bit quantization simultaneously.

\begin{equation}
\label{QuantBits}
\resizebox{.9\hsize}{!}{$
\mathrm{QuantBits}=
\begin{cases}
4& \mathrm{KL}(P,Q_\mathrm{INT4})/\mathrm{KL}(P,Q_\mathrm{INT8}) < \gamma\\
8& \mathrm{KL}(P,Q_\mathrm{INT4})/\mathrm{KL}(P,Q_\mathrm{INT8}) \geq \gamma 
\end{cases}
$}
\end{equation}
%
%
Once the truncation threshold and quantization bit width are determined, a linear layer is subjected to the quantization framework depicted in Figure \ref{overflow}. Our framework notably incorporates flat tensors, which significantly contribute to achieving high-precision per-tensor quantization.









%% file: Experiment.tex

\subsection{Settings}
\paragraph{Baselines.}
Our algorithm is designed with a specific focus on improving inference efficiency in scenarios where per-tensor quantization offers advantages, especially in compute-bound situations. To comprehensively assess its performance, we compared it with two baseline methods: the naive quantization of W8A8 and the SmoothQuant approach. The FlattenQuant method introduces a series of quantization levels, namely O1, O2 and O3, which progressively increase in aggressiveness and efficiency. For more detailed insights into the specific quantization schemes employed in both the baselines and FlattenQuant, please refer to Table \ref{QuantizationSetting}.

\begin{table}[!ht]
\centering
\caption{Quantization setting of the baselines and FlattenQuant. All weight and activations use per-tensor static quantization. In the case of FlattenQuant-O3, we adopt per-tensor quantization instead of per-channel quantization, during the GPTQ optimization process for weight.}\label{QuantizationSetting}

\resizebox{\linewidth}{!}
{
\begin{tabular}{ccc}
\toprule
Method& Bits & Additional Optimizations\\
\midrule
W8A8 & 8& N/A\\
Smoothquant &8& N/A\\
FlattenQuant-O1 &8& N/A\\
FlattenQuant-O2 &4,8 mixed& N/A\\
FlattenQuant-O3 &4,8 mixed& GPTQ\\
\bottomrule
\end{tabular}
}
\end{table}

\begin{table*}[!ht]
\centering
\caption {We evaluate the performance on 5 zero-shot benchmarks (by reporting the average accuracy) and 1 language modeling benchmark (perplexity). The accuracy degradation observed in the FlattenQuant-O1 setting is minimal, as the utilization of 8-bit quantization ensures the preservation of accuracy. Conversely, the O2 setting exhibits a more noticeable loss of accuracy. However, the introduction of GPTQ in the O3 setting compensates for the accuracy loss caused by INT4 quantization. }\label{tab:Accuracy}

\resizebox{\linewidth}{!}
{
\begin{tabular}{ccccccccccccc}
\toprule
Task& \multicolumn{6}{c}{OpenBookQA ($\uparrow$)} & \multicolumn{6}{c}{LAMBADA (OpenAI)($\uparrow$)} \\
\midrule
Model& 125M& 1.3B& 6.7B& 13B& 30B& 66B
& 125M& 1.3B& 6.7B& 13B& 30B& 66B\\
\midrule
FP16& 27.80\%& 33.40\%& 37.40\%& 39.00\%& 40.20\%& 40.80\%&
 37.84\%& 57.90\%& 67.66\%& 68.60\%& 71.39\%& 73.92\%\\
W8A8& 27.60\%& 34.20\%& 28.20\%& 25.60\%& 27.80\%& 27.33\%&
 35.53\%& 54.39\%& 13.31\%& 0.05\%& 0.03\%& 0.02\%\\
SmoothQuant& \textbf{27.20\%}& \textbf{33.60\%}& \textbf{37.40\%}& \textbf{38.85\%}& \textbf{39.63\%}& \textbf{39.82\%}&
 \textbf{37.41\%}& \textbf{55.24\%}& \textbf{67.03\%}& \textbf{68.02\%}& \textbf{71.20\%}& \textbf{72.79\%}\\
FlattenQuant-O1& 27.80\%& 33.00\%& 37.80\%& 39.20\%& 39.27\%& 39.87\%&
 36.73\%& 56.47\%& 66.13\%& 67.41\%& 71.22\%& 73.12\%\\
FlattenQuant-O2& 27.11\%& 31.00\%& 35.60\%& 38.11\%& 37.74\%& 36.52\%&
 35.88\%& 55.63\%& 65.47\%& 66.03\%& 68.58\%& 69.91\%\\
FlattenQuant-O3& \textbf{27.60\%} & \textbf{33.22\%}& \textbf{37.01\%}& \textbf{38.71\%}& \textbf{39.11\%}& \textbf{39.46\%}&
 \textbf{36.13\%}& \textbf{56.51\%}& \textbf{66.41\%}& \textbf{67.20\%}& \textbf{70.82\%}& \textbf{72.63\%}\\

\midrule
Task& \multicolumn{6}{c}{PIQA ($\uparrow$)} & \multicolumn{6}{c}{HellaSwag ($\uparrow$)} \\
\midrule
Model& 125M& 1.3B& 6.7B& 13B& 30B& 66B
& 125M& 1.3B& 6.7B& 13B& 30B& 66B\\
\midrule
FP16& 61.53\%& 71.36\%& 76.49\%& 76.87\%& 78.12\%& 79.76\%&
 31.32\%& 53.73\%& 67.18\%& 69.80\%& 72.27\%& 74.88\%\\
W8A8& 61.26\%& 69.85\%& 59.30\%& 52.50\%& 53.15\%& 53.88\%&
 30.92\%& 51.89\%& 39.31\%& 27.44\%& 28.34\%& 28.01\%\\
SmoothQuant& \textbf{61.75\%}& \textbf{71.04\%}& \textbf{76.10\%}& \textbf{76.12\%}& \textbf{77.52\%}& \textbf{79.01\%}&
 \textbf{31.42\%}& \textbf{53.50\%}& \textbf{66.69\%}& \textbf{67.54\%}& \textbf{70.85\%}& \textbf{72.89\%}\\
FlattenQuant-O1& 61.47\%& 70.89\%& 75.78\%& 76.06\%& 78.01\%& 79.34\%&
 31.55\%& 53.09\%& 65.97\%& 68.68\%& 71.31\%& 73.62\%\\
FlattenQuant-O2& 60.77\%& 69.04\%& 74.51\%& 75.89\%& 76.07\%& 76.52\%&
 31.21\%& 50.09\%& 63.45\%& 66.38\%& 68.51\%& 70.11\%\\
FlattenQuant-O3& \textbf{61.20\%}& \textbf{71.02\%}& \textbf{75.43\%}& \textbf{76.31\%}& \textbf{77.58\%}& \textbf{79.07\%}&
 \textbf{31.56\%}& \textbf{53.40\%}& \textbf{66.31\%}& \textbf{68.70\%}& \textbf{71.35\%}& \textbf{73.10\%}\\
\midrule

Task& \multicolumn{6}{c}{WinoGrande ($\uparrow$)} & \multicolumn{6}{c}{WikiText2 ($\downarrow$)} \\
\midrule
Model& 125M& 1.3B& 6.7B& 13B& 30B& 66B
& 125M& 1.3B& 6.7B& 13B& 30B& 66B\\
\midrule
FP16& 50.43\%& 59.19\%& 65.19\%& 65.03\%& 68.42\%& 68.90\%&
 27.65& 14.62& 10.86& 10.13& 9.56& 9.34\\
W8A8& 49.90\%& 58.40\%& 49.56\%& 51.30\%& 50.74\%& 49.72\%&
 30.10& 15.85& 31.08& 4551.63& 2130.69& 3754.28\\
SmoothQuant& \textbf{50.03\%}& \textbf{59.11\%}& \textbf{64.90\%}& \textbf{64.21\%}& \textbf{68.52\%}& \textbf{68.21\%}&
 \textbf{28.88}& \textbf{15.69}& \textbf{11.43}& \textbf{10.99}& \textbf{10.62}& \textbf{10.55}\\
FlattenQuant-O1& 49.01\%& 58.48\%& 64.56\%& 64.24\%& 68.11\%& 68.33\%&
 28.46& 15.22& 11.28& 10.87& 10.34& 10.24\\
FlattenQuant-O2& 48.21\%& 57.85\%& 62.96\%& 63.21\%& 65.47\%& 66.43\%&
 30.86& 16.80& 13.30& 12.26& 11.72& 11.37\\
FlattenQuant-O3& \textbf{50.35\%}& \textbf{58.90\%}& \textbf{64.77\%}& \textbf{64.41\%}& \textbf{68.10\%}& \textbf{68.12\%}&
 \textbf{28.75}& \textbf{15.94}& \textbf{11.68}& \textbf{11.21}& \textbf{10.88}& \textbf{10.77}\\

\bottomrule
\end{tabular}
}
\end{table*}

\paragraph{Models and datasets.}
We evaluated our proposed FlattenQuant on OPT models \cite{zhang2022opt}, using five zero-shot evaluation tasks: OpenBookQA \cite{2018Can}, LAMBADA (OpenAI) \cite{paperno2016lambada}, PIQA \cite{2019PIQA}, HellaSwag \cite{2019HellaSwag}, WinoGrande \cite{sakaguchi2021winogrande}, and one language modeling dataset WikiText \cite{2017Pointer} to validate algorithm settings. We utilized lm-eval-harness \cite{gao2021framework} to evaluate the individual models.
%
%
\paragraph{Implementation.}
We implemented FlattenQuant in PyTorch \cite{paszke2019pytorch} and worked with the HuggingFace \cite{wolf2019huggingface} integrations of the OPT model families. Our experiments were carried out on a server equipped with four A100 GPUs with 80GB of memory. We implemented quantized linear layers and the batched matrix multiplication (BMM) function for INT8 and INT4 based on the CUTLASS INT8 and INT4 GEMM kernels. We simply replace the original floating-point (FP16) linear layers with our INT8 or INT4 quantized linear layers as the quantized model.\\
To smooth channels, $\alpha$ is set to 0.5. Additionally, in order to establish an appropriate truncation threshold, $\beta$ is set to 1.3. Furthermore, for determining the quantization bit width for each layer, we assign a value of $\gamma$ as 1.86. In addition, the final number of channels after the flatten operation is padded to a multiple of 32 to align the matrix multiplication block.
\subsection{Accuracy Results on LLMs}
%
As indicated by the bolded data in Table \ref{tab:Accuracy}, it is evident that the accuracy achieved by FlattenQuant under the O3 configuration is comparable to that of SmoothQuant. The utilization of the flatten operation effectively reduces the maximum value, thereby alleviating quantization challenges. In addition, the integration of GPTQ enables efficient compensation for errors incurred during the weight quantization process. Notably, it is important to highlight that the GPTQ optimization is performed on the flattened weights of each layer.
%
Table \ref{int4_percent} shows the LLMs corresponding setting in our experiment. Across the 6.7b, 13b, 30b, and 66b models of OPT, our methods consistently achieve nearly 50\% layer quantization using INT4. Additionally, the ratio of flattening is predominantly kept within a range of 25\%. This greatly facilitates GPU memory optimization and enhances inference speed.


\begin{table}
\centering
\caption{The proportion of INT4 quantized layers in LLMs, and the average ratio of expanded channels over original channels.}\label{int4_percent}
\resizebox{\linewidth}{!}
{
\begin{tabular}{ccccccc}
\toprule
Model& 125M& 1.3B& 6.7B& 13B& 30B& 66B\\
\midrule
INT4 layers &35.54\%&36.42\%&47.25\%&45.75\%&48.29\%& 47.37\%\\
Flatten ratio &26.21\%&24.62\%&24.41\%&21.50\%&22.42\%& 21.65\%\\
\bottomrule
\end{tabular}
}
\end{table}
%
\begin{table}
\centering
\caption{Comparison of tensor flattening and matrix multiplication latency with different data type. 
The experimental settings involved a batch size of 8, 2048 tokens, 4096 input and output channels, and 1024 tensor expanded channels.
}\label{flatteninglatency}
\resizebox{\linewidth}{!}
{
\begin{tabular}{ccccc}
\toprule
Operator& Flatten& Gemm (FP16)& Gemm (INT8)& Gemm (INT4)\\
\midrule
Latency (ms) &0.19&3.12&2.35&1.57\\ 
\bottomrule
\end{tabular}
}
\end{table}
\subsection{Memory Consumption and Speedup}
We applied INT4 quantization to approximately 50\% of the layers, which further reduced the GPU memory consumption of the inference process compared with the SmoothQuant. The GPU memory consumption is presented in Table \ref{MemoryConsumption}.
It is worth noting that while the inference process for large batch size and long sequence may consume significant memory due to the key-value cache, this specific component was not included in our experiment. The main reason behind this decision is that even with INT8 quantization, the calculations of key-value pairs require a 16-bit representation, while an 8-bit cache is more suitable for industrial scenarios where additional quantization operations are often required. Ensuring accuracy, however, would necessitate a per-group quantization operation, which falls outside the scope of this paper's primary focus.\\
\begin{table*}[!ht]
\centering
\caption{Memory consumption (GB) of LLMs on different batch size and sequence length.}\label{MemoryConsumption}
\scalebox{0.65}
{
\begin{tabular}{ccccccccccc}
\toprule
 &Sequence Length&\multicolumn{3}{c}{512}&\multicolumn{3}{c}{1024}&\multicolumn{3}{c}{2048}\\
\midrule
 &Batch Size&1&8&32&1&8&32&1&8&32\\
\midrule

\multirow{3}{*}{OPT-30b}&
FP16&59.2&60.4&66.2&60.1&62.3&72.3&62.8&73.4&88.6\\
&SmoothQuant&32.3&33.2&38.2&32.6&33.5&40.5&37.2&42.1&48.9\\
&FlattenQuant-O3&\textbf{26.1}&\textbf{26.8}&\textbf{31.7}&\textbf{26.3}&\textbf{27.5}&\textbf{33.5}&\textbf{31.8}&\textbf{36.1}&\textbf{39.8}\\
\midrule

\multirow{3}{*}{OPT-66b}&
FP16&126.3&126.8&133.5&127.2&129.1&140.1&130.6&143.5&162.6\\
&SmoothQuant&67.4&68.2&73.1&67.3&72.2&75.9&69.6&79.9&91.2\\
&FlattenQuant-O3&\textbf{53.7}&\textbf{55.4}&\textbf{60.2}&\textbf{56.2}&\textbf{59.6}&\textbf{60.1}&\textbf{55.4}&\textbf{61.2}&\textbf{70.5}\\
 
\bottomrule
\end{tabular}
}
\end{table*}
%
%
%
\begin{figure*}[!ht]
	\centering        
	{\includegraphics[width=2\columnwidth]{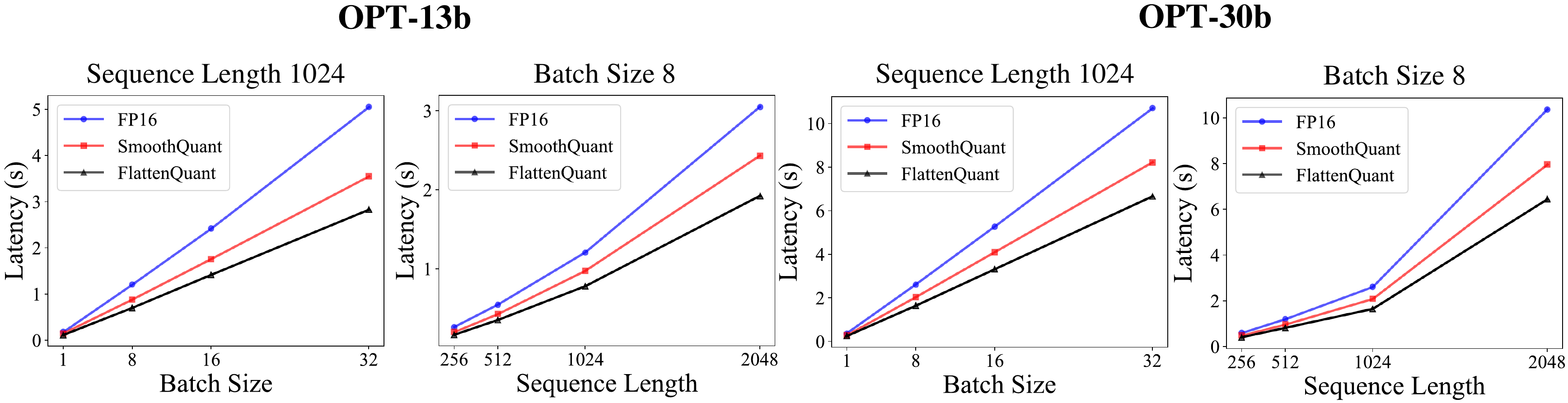}}
    
\caption{
In compute-bound scenarios that involve large batch sizes and long sequences, our per-tensor method replaced FP16 matrix multiplication with INT4 and INT8 alternatives, yielding a considerable improvement in inference speed. This achievement is a direct result of our targeted efforts to overcome compute-bound challenges.
}
\label{Speedfig} 
\end{figure*}
%
Figure \ref{Speedfig} illustrates the inference speed on OPT-13b and OPT-30b for varying batch size and sequence length. The results demonstrate that FlattenQuant achieves nearly double the speed of FP16 and exhibits a significantly accelerated performance compared to SmoothQuant in compute-bound scenarios. When the sequence length remains constant, there is a linear correlation between inference latency and batch size. This correlation arises due to the fact that, once the compute-bound is reached, the inference latency is primarily determined by the data dimension of the matrix calculations. When the batch size is held constant, the theoretical calculation amount increases with the square level of the sequence length, which is also manifested in the inference latency.\\
%
We compared the latency of flatten operation and matrix multiplication, as shown in Table \ref{flatteninglatency}, it can be seen that the latency of tensor flattening operation is very small by comparison.
Hence, the introduction of low-bit calculations through FlattenQuant leads to substantial acceleration, which aligns with the findings depicted in Figure \ref{Speedfig}. 
%
%

\subsection{Ablation Study}
In the FlattenQuant approach, the selection of truncation threshold, quantization bits for each layer (INT4 or INT8), and the presence of channel smooth operator all contribute to the final quantization outcome. In order to determine the optimal quantization process, we conducted a comprehensive ablation study on the WikiText-2 dataset.
\paragraph{Channel smoothing}
Table \ref{SmoothOperatorImpact} shows that the channel smoothing operation brings significant accuracy improvement on three OPT models. We assert that this operation achieves a more uniform distribution of values across activation and weight tensor channels. 
%
%
%
%
\paragraph{Truncation threshold}
In order to strike a balance between quantization precision and resource usage, it is necessary to choose a suitable value for the parameter $\beta$ as shown in Equation \ref{TruncationThreshold}.
The findings obtained from the OPT-6.7 model are shown in Table \ref{truncationThresholdImpact}. When the value of $\beta$ is less than 1.2, the average channel flatten ratio exceeds 30\%, resulting in an increase in GPU memory use. However, the corresponding gain in accuracy is minimal. When the value of $\beta$ exceeds 1.4, there is no substantial change seen in the average channel flatness ratio and GPU memory occupancy. However, a noticeable drop in accuracy becomes evident. Hence, it is necessary to establish the value of $\beta$ within the range of 1.2 to 1.4 in order to get a performance that is well-balanced.
%
\begin{table}[!t]
\centering
\caption{The ablation study to evaluate the performance impact of the channel smoothing.}\label{SmoothOperatorImpact}
\resizebox{\linewidth}{!}
{
\begin{tabular}{ccccccc}
\toprule
Model & \multicolumn{2}{c}{OPT-1.3b}& \multicolumn{2}{c}{OPT-6.7b}& \multicolumn{2}{c}{OPT-13b}\\

Smooth &Yes&No&Yes&No&Yes&No\\ 
\midrule
Wiki PPL ($\downarrow$) &15.94&16.50&11.68&12.16&11.21&11.85\\
\bottomrule

\end{tabular}
}
\end{table}
\begin{table}[!t]
\centering
\caption{
The ablation study on the parameter $\beta$ with respect to accuracy, flatten ratio and GPU memory consumption.}
\label{truncationThresholdImpact}
\resizebox{\linewidth}{!}
{
\begin{tabular}{ccccccc}
\toprule
$\beta$ & 1.1 & 1.2& 1.3& 1.4& 1.5\\
\midrule
Wiki PPL ($\downarrow$) &11.32&11.38&11.68&11.89&12.08\\ 
Flatten ratio &32.25\%&26.85\%&24.41\%&22.01\%&20.56\%\\
Memory (GB) &7.12&6.89&6.75&6.66&6.72\\
\bottomrule
\end{tabular}
}
\end{table}
%
\begin{table}[!t]
\centering
\caption{
Impact of suppressing outlier channels during obtaining truncation threshold.
}\label{OutlierClipImpact}
\resizebox{\linewidth}{!}
{
\begin{tabular}{ccccccc}
\toprule
Model & \multicolumn{2}{c}{OPT-1.3b}& \multicolumn{2}{c}{OPT-6.7b}& \multicolumn{2}{c}{OPT-13b}\\

Outlier clip &Yes&No&Yes&No&Yes&No\\ 
\midrule
Wiki PPL ($\downarrow$) &15.94&16.22&11.68&11.87&11.21&11.74\\
\bottomrule

\end{tabular}
}
\end{table}
\paragraph{Suppressing outlier channels}
In the process of obtaining the truncation threshold, we validated the influence of suppressing outlier channels. Table \ref{OutlierClipImpact} shows that this operation brings significant accuracy improvement on three OPT models. We posit that this operation prevents some outlier channels from excessively affecting the mean of the maximum value of each channel. 
%
\paragraph{Precision selection}
We need to choose an appropriate $\gamma$ (as defined in Equation \ref{QuantBits}) to balance quantization accuracy and resource consumption. Table \ref{QuantizationBitwidthImpact} shows the result on the OPT-6.7 model. When $\gamma$ falls below 1.86, the improvement of accuracy becomes marginal, while the gpu memery occupation increases. When $\gamma$ exceeds 1.88, a noticeable decline in accuracy ensues. Thus, optimal performance can be achieved by setting $\gamma$ between 1.86 and 1.88.

\begin{table}[!ht]
\centering
\caption{$\gamma$ determines the tolerance of the INT4 quantization that affects KL divergence. 
}\label{QuantizationBitwidthImpact}
\resizebox{\linewidth}{!}
{
\begin{tabular}{cccccc}
\toprule
$\gamma$ & 1.82 & 1.84& 1.86& 1.88& 1.90\\
\midrule
INT4 layers &31.21\%&40.85\%&47.25\%&51.67\%&55.85\%\\ 
Wiki PPL ($\downarrow$) &11.35&11.37&11.68&11.95&12.47\\ 
Memory (GB) &7.32&7.02&6.75&6.58&6.29\\
\bottomrule
\end{tabular}
}
\end{table}

%% file: conclusion.tex
We propose FlattenQuant, a per-tensor post-training quantization target for the compute-bound scenarios for large language models, which allows precision lossless in the case of up to 48.29\% linear layer quantized to INT4. FlattenQuant can further reduce inference latency and memory usage compared to SmoothQuant. The use of per-tensor INT4 quantization by FlattenQuant significantly improves inference performance, particularly in scenarios where compute-bound issues arise due to huge batch sizes or long sequence inferences.\\
Furthermore, it is important to underscore some notable constraints: our proposed method is more advantageous for scenarios that tend to compute-bound. It also places certain hardware prerequisites, including the availability of Tensor Cores that poses the capability to handle INT4 data types, hence augmenting the inference efficiency. Additionally, deep operator fusion becomes essential for industrial deployment. By fusing tensor flattening, channel repeat, and subsequent matrix multiplication operators into one single kernel, the resource consumption associated with flatten operations can be further mitigated. Finally, it can be deduced that the impact of our methodology persists as the model expands in size. Consequently, it can be hypothesized that for models beyond 100 billion parameters, we expect to observe comparable improvements in computational efficiency and memory consumption via the implementation of our proposed strategy.


